\documentclass[10pt,twocolumn,letterpaper]{article}
\usepackage{cvpr}
\usepackage{times}
\usepackage{epsfig}
\usepackage{graphicx}
\usepackage{amsmath}
\usepackage{amssymb}
\usepackage{url}
\usepackage{times}
\usepackage{graphicx}
\usepackage{cite}
\usepackage[ruled,lined,slide]{algorithm2e}
\usepackage{algpseudocode}
\usepackage{enumitem}
\usepackage{subfigure}
\usepackage{epstopdf}
\usepackage{lipsum}
\usepackage{multirow}
\usepackage{dsfont}
\usepackage[toc,page]{appendix}
\usepackage{textcomp}

\usepackage[pagebackref=true,breaklinks=true,letterpaper=true,colorlinks,bookmarks=false]{hyperref}

\cvprfinalcopy 

\ifcvprfinal\fi
\begin{document}

\title{DEEP-CARVING : Discovering Visual Attributes by Carving Deep Neural Nets}

\author{Sukrit Shankar$^\dag$, $~~$ Vikas K. Garg$^\ast$  $~~$and$~~$ Roberto Cipolla$^\dag$\\
$^\dag$\fontsize{10.5pt}{11pt}\selectfont Machine Intelligence Lab (MIL), Cambridge University \\ 
$^\ast$\fontsize{10.5pt}{11pt}\selectfont Computer Science \& Artificial Intelligence Lab (CSAIL),  MIT\\
{\tt\small ss965@cam.ac.uk, vgarg@csail.mit.edu, rc10001@cam.ac.uk}
}

\maketitle

\begin{abstract}
\fontsize{9.5pt}{11pt}\selectfont
Most of the approaches for discovering visual attributes in images demand significant supervision, which is cumbersome to obtain. In this paper, we aim to discover visual attributes in a weakly supervised setting that is commonly encountered with contemporary image search engines.

For instance, given a noun (say forest) and its associated attributes (say dense, sunlit, autumn), search engines can now generate many valid images for any attribute-noun pair (dense forests, autumn forests, etc). However, images for an attribute-noun pair do not contain any information about other attributes (like which forests in the autumn are dense too). Thus, a weakly supervised scenario occurs: each of the $M$ attributes corresponds to a class such that a training image in class $m \in \{1,\dots,M\}$ contains a single label that indicates the presence of the $m^{th}$ attribute only. The task is to discover all the attributes present in a test image. 

Deep Convolutional Neural Networks (CNNs) \cite{krizhevsky2012imagenet} have enjoyed remarkable success in vision applications recently.  However, in a weakly supervised scenario, widely used CNN training procedures do not learn a robust model for predicting multiple attribute labels simultaneously. The primary reason is that the attributes highly co-occur within the training data, and unlike objects, do not generally exist as well-defined spatial boundaries within the image. To ameliorate this limitation, we propose \textbf{Deep-Carving}, a novel training procedure with CNNs, that helps the net efficiently carve itself for the task of multiple attribute prediction. During training, the responses of the feature maps are exploited in an ingenious way to provide the net with multiple pseudo-labels (for training images) for subsequent iterations. The process is repeated periodically after a fixed number of iterations, and enables the net carve itself iteratively for efficiently disentangling features.

Additionally, we contribute a noun-adjective pairing inspired \textbf{N}atural \textbf{S}cenes \textbf{A}ttributes \textbf{D}ataset to the research community, \textbf{ CAMIT - NSAD}, containing a number of co-occurring attributes within a noun category. We describe, in detail, salient aspects of this dataset.  Our experiments on CAMIT-NSAD and the SUN Attributes Dataset \cite{patterson2012sun}, with weak supervision, clearly demonstrate that the Deep-Carved CNNs consistently achieve considerable improvement in the precision of attribute prediction over popular baseline methods.

\end{abstract}

\section{Introduction} \label{sec_introduction}
Owing to an exponential increase in the number of images on the web, most image search engines, such as Google, have started resorting to clustering in order to present the search results. In particular, they now categorize the images based on common and key attributes. On receiving a query about \textit{tall buildings}, for instance, Google image search finds thousands of images it thinks contain tall buildings, and then clusters them together into some key attributes such as \textit{night, looking-up}. Analysing the images in these clusters, we observe that the categorization is generally based more on the text information associated with the images than the visual cues. Therefore, the attributes that are missing in the text are rarely inferred in the images. Thus,  it is difficult for the engine to determine which buildings in the cluster of \textit{tall buildings at night} are \textit{curved, glassy, stony}. Hence, the visual cues need to be leveraged for enhancing the search results. 

\textbf{Discovering Visual Attributes under a Practical Scenario - }  Consider a practical system that can predict attribute-specific information within images using visual cues. For simplicity, suppose that we have only 3 attributes of mountains under consideration, viz. \textit{wide-span, hazy and with-reflections}. If we search for \textit{hazy mountains}, it is likely that we get most mountains that seem hazy (with the increased accuracy of search engines), but some of them will also have wide-span, some will exhibit reflections, some will portray both wide-span and reflections, and some neither; however, typically, such information will not be found in the text and thus remain unknown. We can then search for  \textit{wide-span mountains} to get the visual cues (e.g., how a wide-span mountain looks like), but the resulting images might again contain varying and unknown degrees of haziness and reflections. Thus, the following \textbf{problem abstraction} arises naturally while designing a practical system for visual attribute prediction: 

\textit{Each of the $M$ given attributes corresponds to a class. Every training image in the class $m \in \{1,\dots, M\}$ comes with only one label that indicates the presence of the $m^{th}$ attribute. The task is to discover all the attributes present in a test image using this weakly supervised training.} 

\begin{figure*}[!htb]
\begin{center}
   \includegraphics[width=0.99\linewidth, height=0.15\textheight]{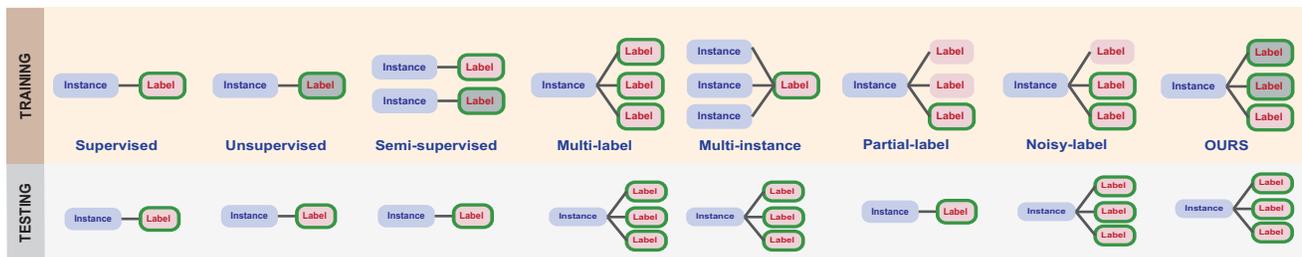}
\end{center}
   \caption{\fontsize{8pt}{9.5pt}\selectfont \textbf{Understanding our Problem Setting for Attribute Prediction: }   \textit{A grey shaded box indicates unavailability. A green outlined box indicates that the label is correct.} \textbf{From Left to Right}: \textbf{Supervised - }Every instance has the correct label available. \textbf{Unsupervised - } No instance is labelled. \textbf{Semi-Supervised - }Some instances have correct labels available, while some are not labelled. \textbf{Multi-label - } An instance has all correct labels available. \textbf{Multi-instance - }Multiple instances together, not individually, have a correct label available. \textbf{Partial-label - } An instance has many labels available, out of which only one can be correct. \textbf{Noisy-label - } An instance has multiple labels available, out of which more than one are possibly correct. \textbf{Our Problem Setting - } An instance can have multiple correct labels, but only a single correct label is available. Also, no negative labels for instances or vice-versa are available. In all cases, the test scenario shows the number of correct labels one needs to predict for a given instance. \textit{Figure is best viewed in color.}}
\label{fig_problem_definition}
\end{figure*}

One might argue that instead of having weak supervision, \textbf{why not search images with multiple attributes under consideration?} Doing a joint search over attributes in the query would lead to an exponential increase in the ambiguity. One might pose another question: \textbf{why not train exhaustively for each attribute with Amazon Mechanical Turk?} Unlike object categories, training for a number of attributes can be prohibitive.\footnote{To see this, note that each attribute is connected to a noun, and with at least 5000 popular noun and adjective synsets each (as per the WordNet \cite{fellbaum1998wordnet}), there will be around 25 million attribute-noun combinations. Typically $10\%$ of such attribute-noun pairs, or roughly 2.5 million, can be deemed to be valid (as per the ImageNet Attribute dataset \cite{russakovsky2012attribute} statistics). Training about 400 images per valid attribute-noun pair will require on the order of $1$ billion positive labels, which is cumbersome to obtain. Alternatively, since a same attribute can exist for many different nouns, one might not have separate classes for noun-attribute pairs; instead one might have an attribute class containing multiple noun categories. Although this decreases the amount of training required, it also increases per-class ambiguity, and usually affects the robustness of the model.}. 

Image datasets for style recognition in scenes \cite{karayev2013recognizing}, object-centric recognition (ImageNet \cite{deng2009imagenet}) and scene-centric recognition (MIT Places \cite{zhou2014learning}) may not provide all the positive correct labels for training instances. Nonetheless, researchers benefit from having to deal with a little mutual overlap across classes in the training set, and requiring to estimate a limited number of (correct) labels in test data.  However, such luxuries do not always extend to the task of attribute prediction, and thus, the weakly supervised setting (as mentioned above in the problem abstraction) is more challenging for predicting attributes than objects/scenes.  


\section{Related Work} \label{sec_related_work}

We now briefly outline the related works on attributes in computer vision, and label prediction in machine learning. Most existing approaches for discovering visual attributes/labels either require significant supervision or have less co-occurrence within the training data, and thus do not conform to our problem setting (see Fig ~\ref{fig_problem_definition} for a succinct overview of related problems). We refer the readers to peruse these works for a holistic overview of the field.

\textbf{Binary Attributes for Better Classification - } 
In the computer vision community,  attribute learning has been conventionally used to provide cues for object and face recognition \cite{lampert2009learning, kumar2009attribute}, zero-shot transfer \cite{lampert2009learning, russakovsky2012attribute}, and part localization \cite{farhadi2009describing, wang2009joint}. There have also been attempts to make learning and classification on categorical attributes robust: for instance, \cite{rastegari2012attribute} strives to make the binary attributes more discriminative on a class basis. However, all these methods require complete attribute labelling for the training images.

\begin{figure*}[!htb]
\begin{center}
   \includegraphics[width=0.52\linewidth, height=0.15\textheight]{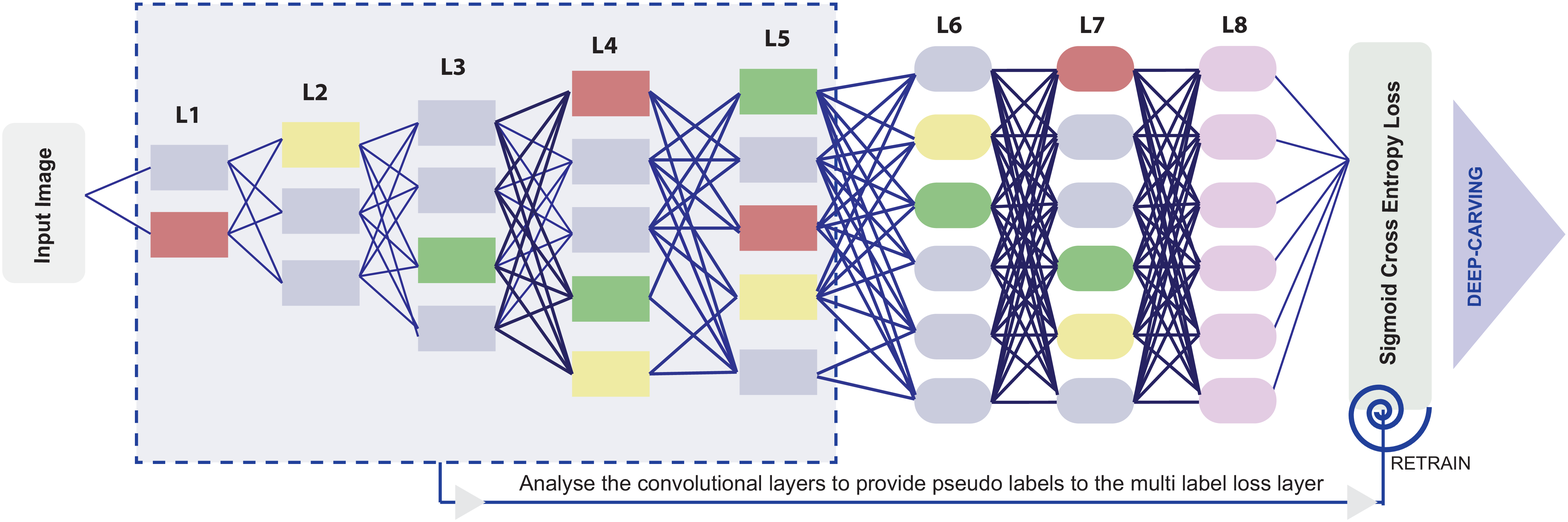}
   \includegraphics[width=0.47\linewidth, height=0.15\textheight]{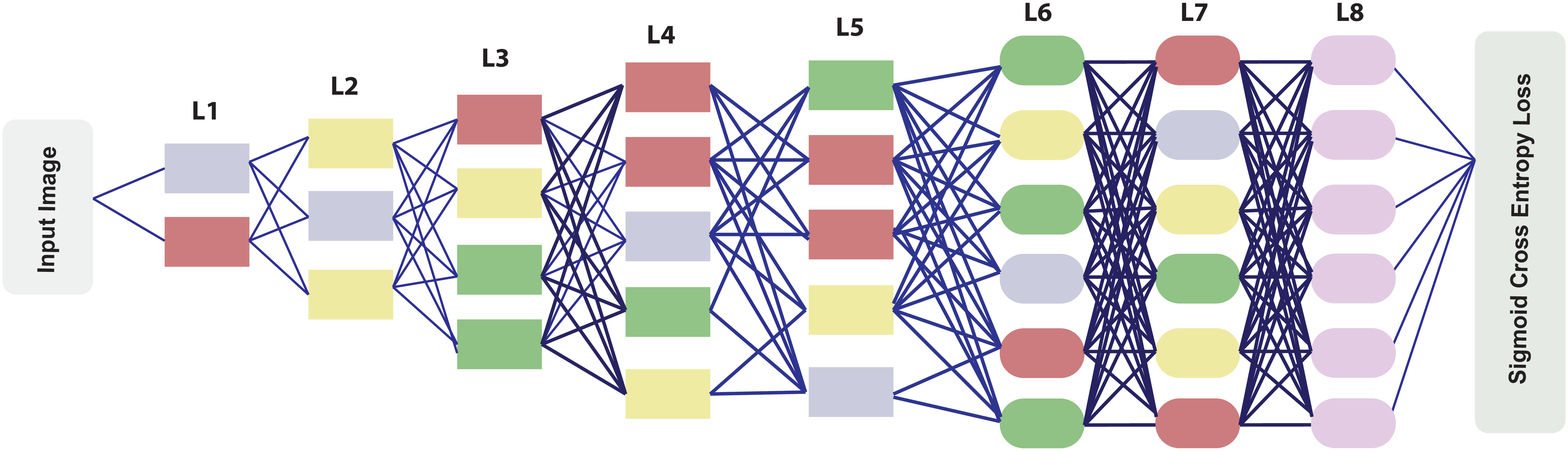}
\end{center}
   \caption{\fontsize{8pt}{9.5pt}\selectfont  \textbf{Illustration of Deep-Carving:  }  
  Deep-carving is a novel training procedure with deep CNNs. During training, the responses of the feature maps are exploited in an ingenious way to provide the net with multiple pseudo-labels (for training images) for subsequent iterations. The process is repeated periodically after a fixed number of iterations once the net has learnt reasonably disentangled feature map representations. This eventually enables the net carve itself iteratively for efficiently predicting multiple attribute labels. \textit{Yellow, Red, Green} coloured feature maps indicate their firing for three attributes in consideration, while \textit{blue} coloured feature maps indicate that they fire for all three attributes. After deep-carving, the feature maps are better disentangled (evaluated through multi-label classification results). The last layer $L8$ is shown in a different color, since it always contains the number of attribute classes as its number of outputs, based on which probabilities are calculated.  \textit{Figure is best viewed in color.}}
\label{fig_deep_carving_illustration}
\end{figure*}

\textbf{Relative Attributes - } Another direction of work \cite{parikh2011relative, shrivastava2012constrained} considers ranking image classes or instances according to the attributes, and training a feature space such that the maximum number of pairwise rank constraints are satisfied. Again, such methods require complete supervision, and thus cannot be applied to our problem. Likewise for the various multi-label ranking methods such as \cite{joachims2002optimizing, chu2005gaussian, bernecker2009probabilistic, soliman2009ranking, guiver2008learning, gong2013deep} considered in the machine learning literature, which propose different types of feature models for efficient rank learning or label prediction, and the associated ensemble methods for multi-label classification such as \cite{rokach2013ensemble, tsoumakas2011random}. Authors in \cite{ma2012unsupervised} try to rank attributes in images in a \textit{completely unsupervised manner}. Their approach behaves rather ambiguously while predicting multiple attributes, and suffers from issues of scalability as well. To counter this problem, \cite{shankar2013semantic} considers a weakly-supervised scenario and estimates the ranking of images based on the attributes.  This approach yields promising results, however, it requires semantic response variables for some images and thus does not apply to our setting. 

\textbf{Predicting Attributes using Textual Information - } Some works like \cite{berg2010automatic, wang2013weakly} aim to estimate the attributes in images, but rely on the availability of text information, which does not hold for our setting. Similarly, \cite{russakovsky2012attribute} tries to predict attributes in the ImageNet dataset \cite{deng2009imagenet}, but  expects all attribute labels to be present in the training data. 

\textbf{Predicting Attributes under Weak Supervision - } The main idea behind \cite{lee2013pseudo, grandvalet2004semi} is to use an Entropy Minimization method to create low-density separation between the features obtained from deep stacked auto-encoders. Their work can be deemed to be nearest to our proposed approach; however, we do not deal with unlabelled data, and tend to follow a more comprehensive approach for attribute prediction. \cite{dengvisual} proposes a weakly supervised graph learning method for visual ranking of attributes, but the graph formulation is heavily dependent on the attribute co-occurrence statistics, which can often be inconsistent in practical scenarios. Researchers in \cite{yu2012weak} attempt to leverage weak attributes in images for better image categorization, but expect all weak attributes in the training data to be labelled. Authors in \cite{cour2011learning} solve the partial labelling problem, where a consideration set of labels is provided for a training image, out of which only one is correct. However, as depicted in Fig ~\ref{fig_problem_definition}, each training image in our problem setting can have more than one correct (but unlabelled) attribute. 

\textbf{Label Prediction with Deep CNNs - } Deep CNNs have recently enjoyed remarkable success for predicting object \cite{deng2009imagenet} and scene labels \cite{zhou2014learning}. Such works contain only one correct label for each training instance, and predict multiple labels for the test images, as in our problem setting (Fig ~\ref{fig_problem_definition}). However, as mentioned before, the same problem setting when applied for attribute prediction is much more challenging, since attributes generally co-occur in abundance within the training instances and cannot be always separated by well-defined spatial boundaries. Thus, deep CNNs clearly require enhancements, more so when false positives also need to be minimized. 

To the best of our knowledge, we are the first to target such a weakly supervised problem scenario for multiple attribute prediction.  We now summarize the \textbf{key contributions of this paper}: 

\small
\begin{enumerate}
\item We emphasize the weakly supervised scenario commonly encountered with image search engines, with an aim to discover multiple visual attributes in test images (see Fig ~\ref{fig_problem_definition}). 

\item We introduce a noun-adjective pairing inspired \textbf{N}atural \textbf{S}cenes \textbf{A}ttributes \textbf{D}ataset (CAMIT-NSAD) having a total of 22 pairs, with each noun category containing a number of co-occurring attributes. In terms of the number of images, the dataset is about three times bigger than the  SUN Attributes dataset \cite{patterson2012sun}.
 
\item We introduce \textbf{Deep-Carving}, a novel training procedure with CNNs, that enables the net efficiently carve itself for the task of multiple attribute prediction. 
\end{enumerate}
\normalsize
\section{Approach} \label{sec_approach}


Recall the problem definition from Section ~\ref{sec_introduction}. Let $\boldsymbol{A} = \{a_1, \ldots, a_M\}$ be the set of $M$ attributes under consideration. We have a weakly supervised training set, $\boldsymbol{S} = \{(\boldsymbol{x_1},y_1), \ldots, (\boldsymbol{x_N}, y_N)\}$ of $N$ images $\boldsymbol{x_1}, \ldots, \boldsymbol{x_N} \in \boldsymbol{X}$ having labels $y_1, \ldots, y_N \in \boldsymbol{A}$ respectively.   Equivalently, segregating the training images based on their label, we obtain $M$ sets $\boldsymbol{S}_m = \boldsymbol{X}_m \times a_m$, where $\boldsymbol{X}_m = \{\boldsymbol{x} \in \boldsymbol{X}| (\boldsymbol{x}, a_m) \in \boldsymbol{S}\}$ denotes the set of $N_m = |\boldsymbol{X}_m|$ images each having the (single) positive training label $a_m, m \in \{1, \ldots, M\}$.  For a test image $\boldsymbol{x_t}$, the task is to predict  $\boldsymbol{y_t} \subseteq \boldsymbol{A}$,  i.e. all the attributes present in $\boldsymbol{x}_t$. 

\textbf{Motivation for Using Deep CNNs to Predict Attributes: } Deep CNNs have recently shown state-of-the-art performance on the tasks of predicting key facial points and facial expressions \cite{sun2013deep, deep_learning_faces}. Although CNNs have been used extensively for object recognition \cite{krizhevsky2012imagenet}, researchers \cite{patterson2012sun} have conventionally used low-level features for attribute prediction in scenes. We compared attribute prediction performances on the SUN Attributes Dataset (with weak supervision) using the state-of-the-art ensemble of low-level features proposed in \cite{patterson2012sun} and the deep CNN architecture proposed in \cite{krizhevsky2012imagenet}, and found that under a weakly supervised scenario, deep CNNs outperformed the low-level features for attribute prediction in scenic images (details provided in Section ~\ref{sec_results}). 

Some researchers have also used Deep Belief Nets (DBNs) \cite{susskind2008generating} for expression (attribute) recognition in faces. However, CNNs are generally more attractive since being translation-invariant, unlike DBNs, they can be used with unconstrained datasets. Though convolutional forms of DBNs exist \cite{lee2009convolutional}, they have not shown much promise over deep CNNs for most of the recognition tasks. Consequently, deep CNNs are an obvious choice to consider for the task of attribute prediction.

\begin{figure}[!htb]
\begin{center}
   \includegraphics[width=0.99\linewidth, height=0.10\textheight]{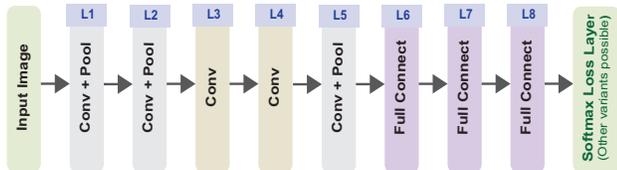}
\end{center}
   \caption{\fontsize{8pt}{9.5pt}\selectfont  \textbf{Block Illustration of AlexNet \cite{krizhevsky2012imagenet}: } The deep convolutional neural net architecture has eight layers ($L1 ~-~L8$) after the input. The last fully connected layer is conventionally followed by a softmax loss layer, but can also be replaced by the likes of Sigmoid Cross Entropy Loss Layer \cite{jia2014caffe}. We use this as the base CNN architecture for all our experiments.}
\label{fig_alexnet}
\end{figure}

\textbf{The CNN Architecture: } Inspired by its huge success \cite{girshick2013rich,jia2014caffe,zeiler2014visualizing}, we use AlexNet \cite{krizhevsky2012imagenet} as our base deep CNN architecture (Fig ~\ref{fig_alexnet}) for all our purposes.  The fully-connected layers have $4096$ neurons each. Max-pooling is done to facilitate translation-invariance. For the fully connected layers, a drop-out \cite{srivastava2014dropout} probability of $0.5$ is used to avoid overfitting. The final fully connected layer  takes the outputs of $L7$ as its input, produces $M$ (equal to the number of classes) outputs through a fully connected architecture, then passes these outputs through a softmax function, and finally applies the negative log likelihood loss. With softmax loss layer, each input image is expected to have only one label. When softmax loss layer is replaced by a sigmoid cross-entropy loss layer, the outputs of $L8$ are applied to a sigmoid function to produce predicted probabilities, using which cross-entropy loss is compute. Here each input can have multiple label probabilities. We refer the reader to \cite{krizhevsky2012imagenet} for details on the kernel and filter sizes of the layers. 



\textbf{Motivation behind Deep-Carving: }	
From our problem description, it is clear that the attribute-specific information needs to be present in a decently segregated form in the output feature vectors. We avail the fact that deep CNNs, even under a weakly supervised scenario, learn a set of reasonably disentangled feature maps during initial stages of training; however, they start to get befuddled (evident from unstable convergence trends) in later stages of training due to lack of all correct labels. 
We thus devise a method to provide the net with pseudo-labels for training images, once the net has initially learned reasonable feature map representations. For this, the responses of the feature maps are analysed in a novel way after every fixed number of iterations during training, and the net eventually carves itself for predicting multiple attribute labels more robustly. 

We call this approach \textit{deep-carving} and argue that it is inherently different from the fine-tuning and dropout procedures. Dropout methods like \cite{srivastava2014dropout} drop parts of the net randomly (without analysing the current training state) to avoid overfitting, while adaptive dropout procedures like \cite{ba2013adaptive} drop parts by analysing the state of the net during training iterations. Fine-tuning procedures \cite{karayev2013recognizing} take a pre-trained net and mainly learn the last layer parameters (while only perturbing the parameters of the other layers) on their training set for a given loss. We instead analyse the net during training to provide a set of new (pseudo) outputs for missing labels in subsequent iterations, which helps the net to carve out attribute-specific feature maps.

\textbf{Training the Net using Deep-carving: }
We consider AlexNet (Fig ~\ref{fig_alexnet}) as our base architecture. With the softmax loss layer, the training of the AlexNet is  typically accomplished by minimizing the following cost or error function (negative log-likelihood):
\small
\begin{equation}
\mathcal{L}_s = -\dfrac{1}{N} \sum_{r=1}^{N}\, \log (\hat{p}_{r,y_r})
\end{equation}
\normalsize

\noindent where the probability $\hat{p}_{r,y_r}, r \in \{1, \ldots, N\}$, is obtained by applying the softmax function to the $M$ outputs of layer $L8$. Letting $l_{r,m}$ denote the $m^{th}$ output for $\boldsymbol{x}_r$, we have
\small
\begin{equation}
\hat{p}_{r,m} = \dfrac{\mathrm{e}^{l_{r,m}}}{\sum_{m'} \mathrm{e}^{l_{r,m'}}}, \qquad ~~m,m' \in \{1, \dots, M\}.
\end{equation}
\normalsize

Note here that for the softmax loss, the labels $y_r \in \boldsymbol{A}$ are encoded in the corresponding range $\{0, \ldots, M-1\}$ for computational purposes. In case one applies the sigmoid cross entropy loss, each image $r$ is expected to be annotated with a vector of label probabilities $\boldsymbol{p}_r$, having length $M$. For our weakly supervised case, the vector $\boldsymbol{p}_r$ is initialized with a very low value of $0.05$ for all images, with $\boldsymbol{p}_r^m = 0.95 ~~~ \forall ~~~ r \in \boldsymbol{X}_m$. With sigmoid cross-entropy loss, the network is trained by minimizing the following loss objective:
\small
\begin{equation}
\mathcal{L}_e =  -\dfrac{1}{N} \sum_{r=1}^{N} \left[\boldsymbol{p_r} \log (\boldsymbol{\hat{p}_r})  ~~ + ~~ (1 - \boldsymbol{p_r}) \log (1 -\boldsymbol{\hat{p}_r})\right]
\end{equation}
\normalsize

\noindent where the probability vector $\boldsymbol{\hat{p}_r}$ is obtained by applying the sigmoid function to each of the $M$ outputs of layer $L8$. 

To learn a deep-carved net, we follow the sigmoid cross-entropy loss since it can take into account the probabilities of multiple labels. For a deep-carving iteration $c$, the following loss is minimized:
\small
\begin{equation}
\mathcal{L}_e^c =  -\dfrac{1}{N} \sum_{r=1}^{N} \left[\boldsymbol{p_r^c} \log (\boldsymbol{\hat{p}_r})  ~~ + ~~ (1 - \boldsymbol{p_r^c}) \log (1 -\boldsymbol{\hat{p}_r})\right]
\end{equation}
\normalsize

\noindent where the probability vector $\boldsymbol{p_r^c}$ is a vector of pseudo-label probabilities (we shall interchangeably refer them as pseudo-labels) computed by Algorithm ~\ref{algo_choosing_features}. 



\begin{algorithm}
\caption{\small{Generating Pseudo-labels for Deep-carving}}
\label{algo_choosing_features}
\footnotesize
{
\For{all feature maps $f$ in convolutional layers} {
 	\For{all attribute classes $a_m \in \boldsymbol{A}$} {
 		\For{all images $r \in \boldsymbol{X_m}$} {
 			Calculate $w_r^m$, average spatial response at $f$ for $r$ 
		}
		Average $\boldsymbol{w}^m$ over $r$ to produce $t_m$ \\
		Assign $h_f^m = t_m$. 
	}
}

$\boldsymbol{h_f}$ is the histogram of average responses at feature map $f$ from all training images for $M$ attribute classes. 

\For{all images $r$ in the training set $\boldsymbol{S}$} {
	\For{all feature maps $f$ in convolutional layers} {
		Calculate $v_f^r$, average spatial response of $r$ at $f$ \\
		\For{all attribute classes $a_m \in \boldsymbol{A}$} {
			\eIf{$a_m == y_r$}{
				$z_r^{f,m} = 0.95$
				}
				{
				\eIf{$\gamma h_f^m \leq v_f^r \leq h_f^m$}{
					$z_r^{f,m} = v_f^r / h_f^m$ \\ 
				}
				{
					$z_r^{f,m} = 0.05$ 
				}
			}
		}
		Average $\boldsymbol{z}_r^m$ over all feature maps $f$ to obtain $\boldsymbol{b_r}$ of length $M$. 
	}
	Form the pseudo labels as $p_r^{c,m} = b_r^m$. Here $c$ stands for the deep-carving iteration. 
}
}
\end{algorithm}  

The method outlined in Algorithm ~\ref{algo_choosing_features} was optimized on the GPU for computational efficiency; however, we have presented the algorithmic steps in a much simpler way to enhance didactic clarity. Note that during the generation of pseudo-labels, we do not change the initially available labels in the training set $\boldsymbol{S}$. The process of predicting pseudo-labels is repeated for each deep-carving iteration $c$, which is chosen periodically after every 5 epochs, once we have already trained for around 60 epochs. Thus, we are delivering the pseudo-labels to the net after some fixed intervals, and that too after the net has initially learnt reasonably disentangled feature maps. 

We only consider the feature maps of convolutional layers for Algorithm ~\ref{algo_choosing_features}. Ideally, the fully connected layers learn their parameters taking the inputs from the convolutional layers and minimizing the cross-entropy loss with original (weakly supervised) labels. After a deep-carving iteration, the net considers the pseudo-labels as its new set of labels for all subsequent iterations till the successive deep-carving iteration. This helps the net to slowly carve itself for efficiently predicting the attribute labels. For all our experiments, we set $\gamma = 0.7$; this is empirically selected and indicates that pseudo-labels are only assigned when the chances of co-occurrence of the missing attributes are significantly high. 

For a given deep-carving iteration $c$, the pseudo-label probabilities generated by Algorithm ~\ref{algo_choosing_features} are different from the output probabilities that the net would have generated. This is because the latter is affected by the fully connected layer parameters that are learnt based on weakly supervised label set, unlike the former. 

\textbf{Prediction using a Learned Model: }
Given a test image, the number of positive labels (say $K$) is known from the ground-truth. Thus, $K$ denotes the number of correct attributes that need to be predicted for the respective test image. Let $\boldsymbol{T}$ contain the positive labels for the test image. Given the sorted (in descending order) probabilities for the test image from the prediction model, we pick top $K$ predictions. Let the set $\boldsymbol{P}$ contain these predicted labels. Both $\boldsymbol{T}$ and $\boldsymbol{P}$ have cardinality $K$. We then calculate the number of true and false positives using $\boldsymbol{T}$ and $\boldsymbol{P}$, and use precision as our performance metric. Note that this is a stricter performance metric compared to the conventional top-$K$ accuracy, where the presence of at least one correct label out of the top $K$ predictions suffices. We thus believe that our chosen performance metric helps better gauge the prediction models. 

\section{Results and Discussion} \label{sec_results}

We now present the results of our experiments with deep-carved CNNs and several baselines on two natural scenes attribute datasets. We also provide details on and motivation behind the new attribute database we introduce in this paper. 

\textbf{Types of Visual Attributes: } In the computer vision literature, many different categories for visual attributes have been considered, with the most common being: (a) Shape \textit{(round, rectangular, etc.)}, (b) Texture \textit{(wet, vegetation, shiny, etc.)}, (c) Proper Adjectives \textit{(cute, dense, chubby, etc.)}, (d) Nouns that cannot be regarded as objects or parts (resorts, sunset, etc.), (e) Colour \textit{(red, green, grey)}, (f) Nouns that denote objects or specific parts of an object \textit{(oceans, flowers, clouds, etc.)}, and (g) Verbs that define a human body pose or activity \textit{(hiking, farming)}. Most of these attribute categories are covered in the SUN Attributes Dataset \cite{patterson2012sun}. In this paper, we only consider attributes that fall into the categories \textit{(a),(b),(c)}, or \textit{(d)}. We make this choice to ensure that we do not try to solve a problem under the paradigm of attribute prediction that can be solved more efficiently using existing approaches in computer vision. For instance, color attributes can generally be discovered by color histograms, human activities (functions) are amenable to pose or activity recognition methods, and nouns that refer to objects can be predicted by using large-scale object recognition datasets (like ImageNet).

\begin{figure*}[!htb]
\begin{center}
   \includegraphics[width=0.85\linewidth, height=0.12\textheight]{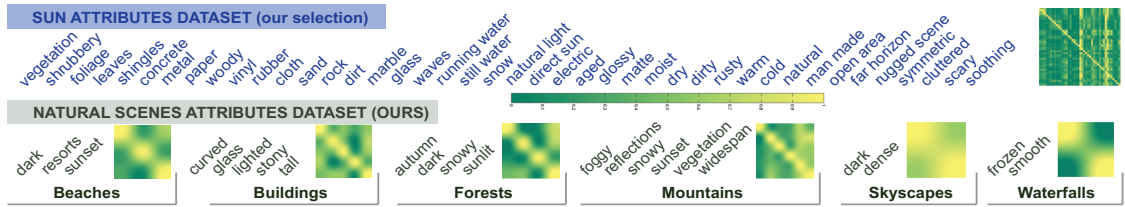}
\end{center}
   \caption{\fontsize{8pt}{9.5pt}\selectfont  \textbf{Attribute Choices and Co-occurrence:  } The figure shows the attributes considered for SAD and CAMIT-NSAD. SAD contains 42 attributes and CAMIT-NSAD contains 18 attributes and 22 attribute-noun pairs. Note that the images in SAD contain attributes like \textit{marble, glass, sand, cloth, etc.} as textures, instead of object-like things. For each dataset, attribute co-occurrence matrices are shown. Each matrix is square, with rows and columns corresponding to the respective attributes in the order in which they are written. Thus, for SAD, matrix is of size $42 \times 42$, and so on.  Let the set of images that contain the attribute represented by a given row be $\boldsymbol{C}$. Then, each column entry in that row is the number of images from $\boldsymbol{C}$ that contain the attribute represented by that column divided by the total number of images in $\boldsymbol{C}$. Thus, diagonal elements are always one, and the co-occurrence statistics is contained in off-diagonal elements. A yellowish pixel indicates greater co-occurrence than a green one. CAMIT-NSAD generally shows high co-occurrence within a noun class as compared to SAD. However, models generally benefit from less co-occurrence of nouns. For some mutually exclusive attributes such as \textit{frozen} and \textit{smooth} for waterfalls, there is no co-occurrence and thus the off-diagonal elements are all green. The co-occurrence statistics are known for the test data sets, and not training, since complete annotations are available only for the test images. Since test set is taken from the same pool of images as that of the training set, co-occurrence statistics presented for test can be deemed to be roughly the same for training data as well. The matrices have been scaled appropriately for better visibility. \textit{Figure is best viewed in color.}}
\label{fig_attribute_choices}
\end{figure*}

\textbf{SUN Attributes Dataset \cite{patterson2012sun}: } The SUN Attributes dataset (SAD) has 102 attributes and contains a total of 14,340 images depicting natural scenes. Each image has annotations to indicate the degree of presence of all 102 attributes. Each positive label in SAD is associated with a confidence value. Confidence values of $0.66$ and $1$ suggest \textit{strong} presence of an attribute, while a confidence value of $0.33$ indicates an ambiguous presence. Given the types of attributes that we ought to consider in this paper, we select 42 suitable attributes (listed in Fig ~\ref{fig_attribute_choices}) out of 102 choices. 

To use  SAD for our weakly supervised scenario, for each attribute class, we algorithmically choose images from SAD that have a \textit{strong} presence of that attribute. We choose at least 250 images for each attribute class, while ensuring that the number of overlapping images across attribute classes is minimal. We thus obtain 22,084 images for  training, 3056 images for validation and 5618 images for testing. The training set contains at least 150 images for each attribute class. The training and validation images chosen for a particular attribute class are all given a single label that indicates the presence of the respective attribute. For each test image, the ground truth comprises of possibly multiple positive labels, thereby indicating \textit{strong} presence of multiple attributes. 

Note that we choose the images from SAD with a slight overlap of images across attribute classes, without introducing any fundamental change to our problem setting, to aptly capture the common real world scenario: in image search engines, there is a small possibility of obtaining same images for different attribute-related queries. For instance, we expect some overlap in the results retrieved from queries for \textit{sunset beaches} and \textit{resort beaches}, since some images in the two collections might have both \textit{beaches} and \textit{resorts} in their textual information.  We preferred SAD over other attribute datasets such as ImageNet  \cite{russakovsky2012attribute} and OSR  \cite{oliva2001modeling} since these contain few attributes of our interest. Also, we do not consider style recognition datasets like Aesthetic Visual Analysis \cite{murray2012ava} in this paper, since they mainly contain photographic attributes instead of general scene attributes. However, our algorithm is generic enough to be applied to style recognition datasets as well.  

\begin{figure*}[!htb]
\begin{center}
   \includegraphics[width=0.90\linewidth, height=0.12\textheight]{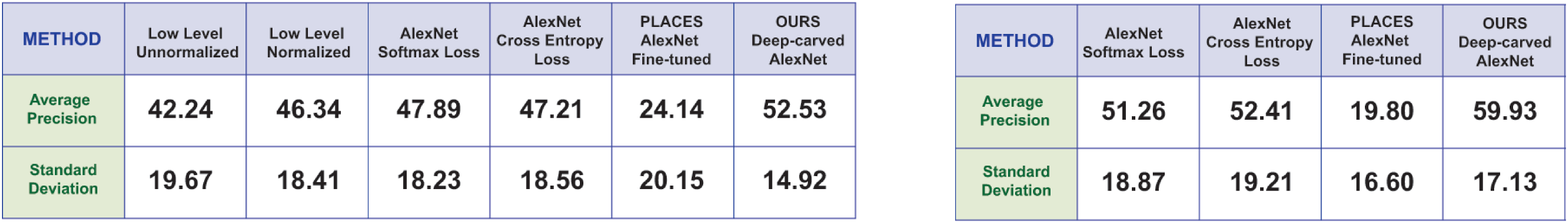}
\end{center}
   \caption{\fontsize{8pt}{9.5pt}\selectfont  \textbf{Comparison of Attribute Prediction Results:} \textit{Left Table - } Average precision on the SAD (weakly supervised) with combined-low-level features, normalized combined-low-level features and AlexNet (Fig ~\ref{fig_alexnet}) with Softmax and Sigmoid Cross Entropy Loss Layers. AlexNet outperforms the low-level feature methods. \textit{Right Table - } Average precision on CAMIT-NSAD for AlexNet with Softmax and Sigmoid Cross Entropy Layers. In both cases, fine-tuned AlexNet over the MIT Places dataset does not perform well, while deep-carved nets exhibit significant improvement over the AlexNet baselines.}
\label{fig_results_comparison}
\end{figure*}

\textbf{Natural Scenes Attributes Dataset: }
SAD contains attributes for natural scenes in general. However, it does not segregate attributes in relation to a specific noun. In practice, people typically search for an attribute-noun pair rather than an attribute. For instance, it is more common to search for \textit{beautiful valleys} instead of just \textit{beautiful}. Therefore, we introduce the Cambridge-MIT \textbf{N}atural \textbf{S}cenes \textbf{A}ttributes \textbf{D}ataset (CAMIT-NSAD) that contains attribute-noun pairs. For a given noun, the attributes co-occur significantly in CAMIT-NSAD. Moreover, different nouns can co-occur in a scene occasionally  (\textit{dark \textbf{skyscapes} with sunset \textbf{beaches}, etc.}). Some of the most popular attributes and nouns on 500px / Flickr have been selected for CAMIT-NSAD (refer to Fig ~\ref{fig_attribute_choices} for a complete overview). 

CAMIT-NSAD contains 46,008 training images, with at least 500 images for each attribute-noun pair. The validation set and the test set contain  2104 and 2967 images respectively. All images in CAMIT-NSAD were collected from 500px, Flickr and Google Search engine, and manually cleaned for every attribute-noun pair. For ground truth, the test set images were annotated for the presence/absence of attribute-noun pairs.   CAMIT-NSAD, as a natural scenes attributes dataset, is quite different to SAD. While the noun-attribute pairs make object and attribute detection more specifically related, there is generally lesser co-occurrence across classes, but much more within a noun class. Although this helps to make the prediction model robust, discovering attribute-noun pairs still remains challenging with deep CNNs.

\begin{figure*}[!htb]
\begin{center}
   \includegraphics[width=0.90\linewidth, height=0.18\textheight]{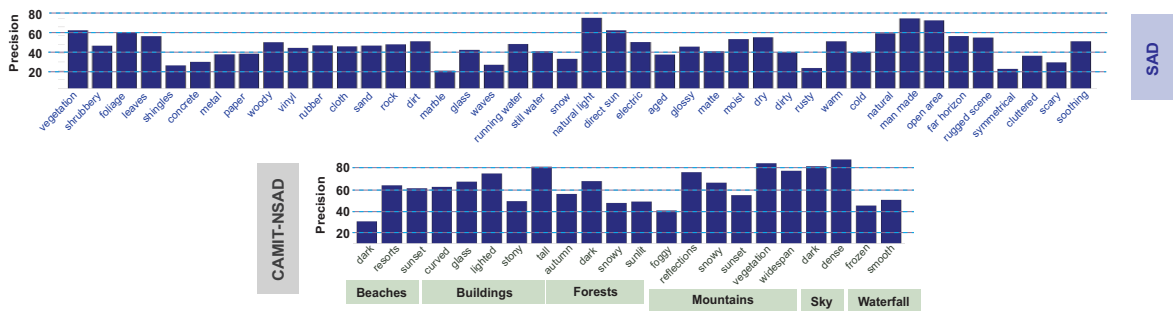}
\end{center}
   \caption{\fontsize{8pt}{9.5pt}\selectfont  \textbf{Attribute-Wise Results with our Deep-Carved CNNs:} The precision of predicting attributes / attribute-noun pairs with deep-carved nets for SAD and CAMIT-NSAD is bar-plotted. It can be seen that the attributes that are less abstract and have lesser chances to co-occur with other attributes in an image are easily predicted in general. Attributes such as \textit{symmetrical} which involve structural relations are difficult to predict, unless they are paired with a specific noun category (\textit{mountains-reflections}). Attributes such as \textit{dark} beaches can be sometimes ambiguous for the net, since evening and night beach images are both considered as dark; however, their color tones are different.}
\label{fig_results_attributewise}
\end{figure*}

All images in SAD and CAMIT-NSAD are $256 \times 256$ RGB. For a test image, the number of positive labels (say $K$) was known from the ground-truth, using which precision was calculated according to Section ~\ref{sec_approach} for gauging the performance. 

All our deep learning related experiments were conducted on NVIDIA TITAN GPUs using Caffe Library \cite{jia2014caffe}. We configured Caffe to use Stochastic Gradient Descent (SGD), and stopped the training after a maximum of 500 epochs. Manual tuning procedures with SGD were carried out using the heuristics mentioned in \cite{jia2014caffe} and \cite{bottou2012stochastic}.

\textbf{Low-level Features vs Deep CNNs for Attribute Prediction on SAD: }
We compared attribute prediction using deep CNNs, on SAD (with weak supervision), with the state-of-the-art low-level feature ensemble of \cite{patterson2012sun}, which combines color histograms, histogram of oriented gradients \cite{dalal2005histograms}, self-similarity, and gist descriptors \cite{oliva2001modeling}. We tried two cases with combined low-level features. First, when the features were simply concatenated; and second,  when the features were individually normalized before concatenation. Note that unlike \cite{patterson2012sun}, we did not learn separate classifiers for each low-level feature, in order to draw a fair comparison with deep net features. As shown in Fig ~\ref{fig_results_comparison}, AlexNet performed better than the low-level features. Normalized low-level feature combination significantly outperformed the simply concatenated one\footnote{Some researchers \cite{karayev2013recognizing} have tried to concatenate the low-level features and deep net features to improve results for the recognition of styles in scenes. However, we do not follow this approach, since the main aim of our work is to show that deep-carving can help improve the conventional deep learning results on attribute recognition in a weakly supervised scenario.}. 

\begin{figure*}[!htb]
\begin{center}
   \includegraphics[width=0.99\linewidth, height=0.14\textheight]{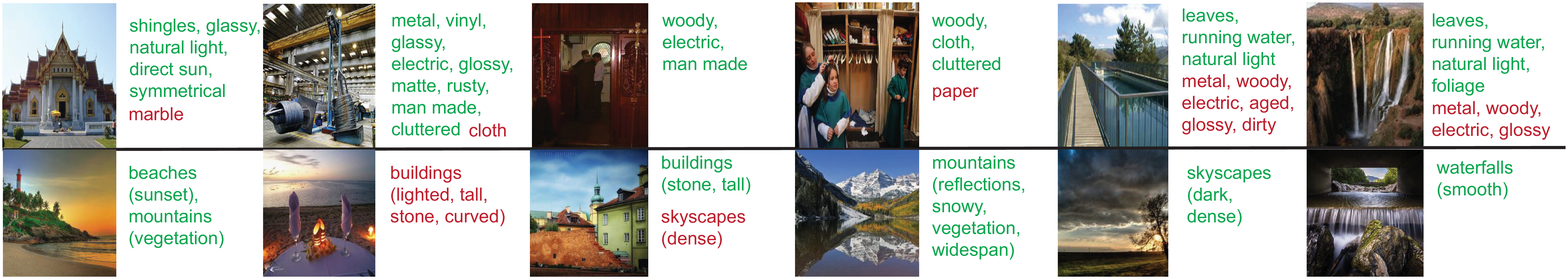}
\end{center}
   \caption{\fontsize{8pt}{9.5pt}\selectfont  \textbf{Attribute Predictions with our Deep-Carved CNNs:} The correctly predicted attributes (true positives) shown in green, and the wrongly predicted ones (false positives) shown in red for various instances in SAD (top row) and CAMIT-NSAD (bottom row) with our deep-carved CNNs. The attributes that are abstract in nature or heavily co-occur with other attributes, are generally predicted with lesser accuracy. \textit{ Figure is best viewed in color.} }
\label{fig_all_results_end}
\end{figure*}

\begin{figure*}[!htb]
\begin{center}
   \includegraphics[width=0.75\linewidth, height=0.12\textheight]{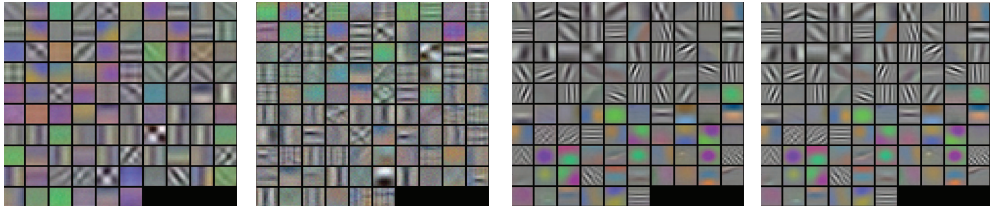}
\end{center}
   \caption{\fontsize{8pt}{9.5pt}\selectfont  \textbf{Visualization of the Filters for the first Convolutional Layer of:} \textit{From Left to Right - } Deep-carved AlexNet for SAD and CAMIT-NSAD, fine-tuned AlexNet over Places205 Model (MIT Places Dataset) for SAD and CAMIT-NSAD. There are 96 filters of sizes $11 \times 11$ with 3 channels for each learnt model, and are shown here on a $10 \times 10$ grid. There is hardly any difference between the last two filter sets, since during fine-tuning, the last full connected layer parameters were only learnt from scratch (random initialization), while the parameters of all the other layers were only perturbed. This is the standard outline for fine-tuning pre-trained Caffe Models as listed out in \cite{jia2014caffe}. \textit{ Figure is best viewed in color.} }
\label{fig_net_conv1_features}
\end{figure*}

\textbf{Baselines: } We consider three major baselines for comparing our deep-carved nets. First, we choose Alexnet with softmax loss layer because of its immense popularity and success in  vision recognition tasks. Second, we choose AlexNet architecture with a sigmoid cross-entropy loss layer, since it better mimics the multi-label prediction scenario as compared to a softmax loss layer. Third, owing to the recent success of MIT Places dataset for scene recognition, we fine-tune their pre-trained model with our training data using the softmax layer loss. Their pre-trained models follow the AlexNet architecture, and during fine-tuning, we mainly learn the $L8$ layer parameters while allowing the parameters of other layers only to get perturbed\footnote{This is done in Caffe by setting the \texttt{blobs\textunderscore lr} parameter to 10 for layer $L8$, while keeping it 1 for the other layers. The number of outputs in $L8$ are also changed to the number of attribute classes $M$.}.

\textbf{Comparison with Deep-carved CNNs: } Fig ~\ref{fig_results_comparison} shows a comparison of the baselines with our deep-carved nets on SAD and CAMIT-NSAD, while Fig ~\ref{fig_results_attributewise} shows attribute-wise performance of our deep-carved CNNs.  It is clear that deep-carved CNNs significantly outperform the baselines. Note that the results with fine-tuning of Places models for our datasets show drastically decreased performance. Although the MIT Places dataset (we consider the 205 categories variant) contain similar images to that of SAD and CAMIT-NSAD, the fine-tuned net mostly outputs low probabilities for the correct attributes, as it gets confused having been apriori trained on a lot of scene categories. The results might get better if one fine-tunes with more number of layers instead of just $L8$.  Fig ~\ref{fig_net_conv1_features} helps to understand this better. It can be seen that the fine-tuned models generally contain very crisp object-specific (edge-like) filters in their first convolutional layer, and  seem less oriented towards learning attribute-specific filters (color patterns, mixed textures). On the other hand, deep-carved nets for CAMIT-NSAD learn some object-specific and some attribute-specific filters. This is understandable since the classes in training set of CAMIT-NSAD contain noun-attribute pairs. The deep-carved nets on SAD learn very less of object-specific filters and more of color patterns, as the training classes are not particular to any noun category, rather contain multiple noun categories. One might infer the merging color patterns to represent scene-specific features; however, our experiments on CAMIT-NSAD show that such patterns more precisely encode attributes of well-categorized scenes. Although inversion of CNN features \cite{simonyan2013deep} for different input images might be more appropriate for analysing the filters and feature map responses, the marked differences in the filters of the first convolutional layer give a fair indication of how the net might be getting biased. 

Fig ~\ref{fig_all_results_end} shows examples of the attributes correctly / incorrectly detected for test images in SAD and CAMIT-NSAD. When the co-occurring attributes are abstract and heavily co-occur with other attributes within an image, the number of false positives generally increases. Attribute-wise accuracy with deep-carved nets can be seen in Fig ~\ref{fig_results_attributewise}.

\section{Conclusions and Future Work} \label{sec_conclusions}
We have targeted  the weakly supervised scenario commonly encountered with image search engines, with an aim to discover multiple visual attributes in images.  We have proposed a novel training procedure with CNNs called Deep-Carving, that helps the net efficiently carve itself for the task of multiple attribute prediction. We have also introduced a noun-adjective pairing inspired natural scenes attributes dataset (CAMIT-NSAD), with each noun category containing a number of co-occurring attributes. Our results show that deep-carving significantly outperforms several popular baselines for our weakly supervised problem setting. CAMIT-NSAD and the pre-trained deep-carved Caffe Models can be accessed from \texttt{http://mi.eng.cam.ac.uk/$\sim$ss965/}.



{\small
\bibliographystyle{ieee}
\bibliography{mainBib}
}

\end{document}